\documentclass[]{spie} 

\usepackage[]{graphicx}
\usepackage{epstopdf}
\usepackage{float}
\usepackage{bm}
\usepackage{amsmath}
\usepackage{amsfonts}
\usepackage{amssymb}
\usepackage{algorithmic}
\usepackage{algorithm}
\usepackage{color}
\usepackage{eso-pic}

\title{Adaptive coherence estimator (ACE) for explosive hazard detection using wideband electromagnetic induction (WEMI)}
\author{Brendan Alvey, Alina Zare, Matthew Cook, Dominic K. C. Ho  \\ University of Missouri, Columbia, MO 65211}
\begin{document}

\AddToShipoutPictureBG*{
  \AtPageUpperLeft{
    \hspace{\paperwidth}
    \raisebox{-\baselineskip}{
      \makebox[0pt][r]{Distribution Statement A: Approved for public release: distribution is unlimited \hspace{1cm}}
}}}

\maketitle

	\begin{abstract}
The adaptive coherence estimator (ACE) estimates the squared cosine of the angle between a known target vector and a sample vector in a transformed coordinate space. The space is transformed according to an estimation of the background statistics, which directly effects the performance of the statistic as a target detector. In this paper, the ACE detection statistic is used to detect buried explosive hazards with data from a Wideband Electromagnetic Induction (WEMI) sensor. Target signatures are based on a dictionary defined using a Discrete Spectrum of Relaxation Frequencies (DSRF) model. Results are summarized as a receiver operator curve (ROC) and compared to other leading methods. 
	\end{abstract}


	\section{Introduction} \label{intro_lbl}
A prototype wideband electromagnetic induction (WEMI) sensor has been developed and investigated in the literature for buried explosive object detection and discrimination \cite{WEMI_0, WEMI_1, WEMI_2, WEMI_3, WEMI_4, WEMI_5, JOMP, DSRF}. In this paper, the ACE detector is proposed and evaluated for the detection of buried explosive objects given data from this WEMI sensor.  

		\subsection{Background on the WEMI Sensor} \label{bckg_lbl}
The sensor used in this investigation emits energy, via a transmit coil, in the form of a time varying electromagnetic field. This field causes elements below the sensor to induce their own electromagnetic field, which is then picked up by one or more receive coils on the sensor. The sensor operates at twenty-one logarithmically spaced frequencies. Each sample collected by this sensor is stored as a twenty-one dimensional complex vector which represents the measured response at each of the operating frequencies. The WEMI sensor is attached to a cart, which also has GPS sensors attached to it to record the UTM spatial coordinates corresponding to each sample. 

The data measured by the sensor is filtered in the down-track direction by convolving it with a zero-mean sine filter, as described in (Scott, 2008)\cite{WEMI_0}. As described by Scott, this filtering has at least four benefits. First, the ground response is attenuated by differencing nearby sections of ground. For similar reasons, the drift in the system is also mostly removed by this filtering. In addition, the filter has the effect of averaging nearby points which increases the signal to noise ratio. Lastly, the filtered data has a maximum response directly over targets, rather than a minimum due to the geometry of the sensor. 

In addition to filtering, the data is normalized before a detection statistic is computed. Each sample is extended to a forty-two dimensional vector by concatenating the real and imaginary responses. The real mean is subtracted and each sample is divided by its L2 norm. This results in each data sample having unit length and zero real mean so that scale and real shift variations may be ignored. In this paper, the detection algorithms investigated use a dictionary of target signatures.  These dictionaries undergo an identical normalization, prior to application.

		\subsection{Discrete Spectrum of Relaxation Frequencies Dictionary} \label{dsrf_lbl}
\begin{figure}[thb]
				\centering
				\includegraphics[width=0.45\textwidth]{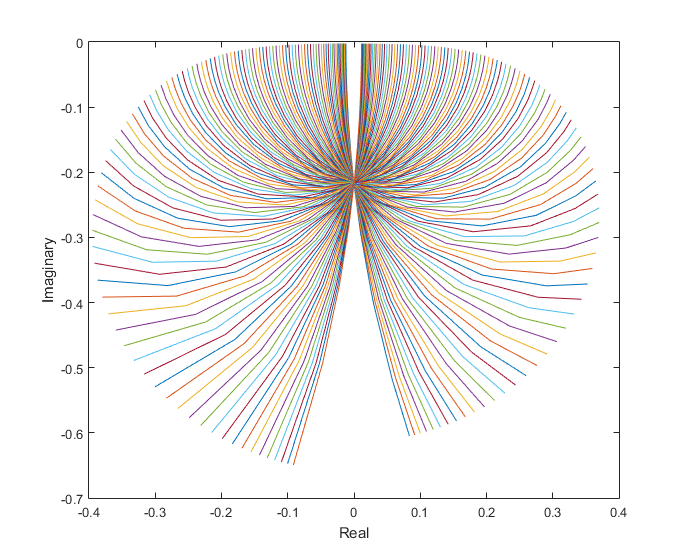}
				\caption{Discrete Spetrum of Relaxation Frequencies (DSRF) dictionary.}
				\label{dsrf_dict_fig}
			\end{figure}

Detection algorithms that leverage a dictionary of target signatures based on the Discrete Spectrum of Relaxation Frequencies (DSRF) have been shown repeatedly in the literature to be useful for buried explosive object detection using this WEMI sensor \cite{DSRF, JOMP, WEMI_2}. This dictionary is generated from a model of the electromagnetic induction (EMI) response of a target \cite{DSRF_0}. This EMI frequency response is given by Equation \ref{dsrf_freq_resp_eq},
			\begin{equation} \label{dsrf_freq_resp_eq}
				H(\omega) = c_0 + \sum_{k=1}^{L} \frac{c_k}{1 + \frac{j\omega}{\zeta_k}}
			\end{equation}			
where $c_0$ is the shift, $L$ is the model order, $c_k$ is the real spectral amplitude and $\zeta_k$ is the relaxation frequency. One hundred dictionary elements are generated using relaxation frequencies logarithmically spaced from 45Hz to 670Khz to model metallic objects. The shift is set to zero, as the real mean is relatively consistent amongst buried objects, and thus is removed without losing much useful information. For our experiments, we populate a dictionary of one hundred elements using the range of operating frequencies used by the WEMI sensor. This dictionary is shown in Fig. \ref{dsrf_dict_fig}.

		\subsection{Joint Orthogonal Matching Pursuits}
One method which has been applied to WEMI data for the task of landmine detection previously is Joint Orthogonal Matching Pursuits (JOMP) \cite{JOMP, WEMI_2}. JOMP is an extension of the previously formulated OMP \cite{OMP}, which compares a single sample at a time to a target signature. With OMP, a sample, $x_j$, is modeled as a sparse linear combination of dictionary elements, $\bm{D} = \left\{{\bm{d}_1, \bm{d}_2, ..., \bm{d}_M}\right\}$, described by $\bm{x}_j = \Sigma^m_{k = 1}w_{kj}\bm{d}_k$ where $m<<M$. JOMP extends OMP to compare multiple samples simultaneously.  In our application, two samples symmetrically spaced from a central point are considered jointly at one time. This spacing is set based on the sample-rate and average velocity of the sensor as it moves down-track. The confidence value for JOMP is given by Equation \ref{JOMP_eq},
			\begin{equation} \label{JOMP_eq}
				c_{\bm{x}} = \frac{1}{1 + \frac{1}{n} \Sigma_j r_j}
			\end{equation}	
where $r_j$ is the residual error for the $j^{th}$ sample being considered where $n$ is the total number of samples being considered simultaneously.  The residual error is defined as $r_j = \|\bm{y} -\bm{x}\|_2^2$ where $\bm{y}$ is an original data sample and $\bm{x}$ is the OMP reconstruction of $\bm{y}$, defined above. This gives a measure of how well the dictionary can represent the points being considered. JOMP is more resilient to random noise than OMP because both points must match well to the same target signature to return a high JOMP confidence.
			
\section{The Adaptive Coherence Estimator} \label{ace_lbl}
The Adaptive Coherence Estimator (ACE) has been shown to be an effective detection statistic \cite{ACE_UMP, ACE_2, ACE_3, ACE_4, ACE_5}.ACE assumes that a target data sample can be modeled as a linear combination of a known target signature and random Gaussian noise. The WEMI sensor being investigated is likely to return data that closely matches this model. ACE uses an estimate of the background mean and background covariance to transform the coordinate space before comparing a known target vector to a data sample. The standard formulation of the ACE detection statistic is shown in Equation \eqref{ace_eq},
			\begin{equation} \label{ace_eq}
				ace(\bm{x}) = \frac{[(\bm{t} - \bm{\mu})^T\bm{\Sigma}^{-1}(\bm{x} - \bm{\mu})]^2}{[(\bm{t} - \bm{\mu})^T\bm{\Sigma}^{-1}(\bm{t} - \bm{\mu})][(\bm{x} - \bm{\mu})^T\bm{\Sigma}^{-1}(\bm{x} - \bm{\mu})]}
			\end{equation}
where $\bm{t}$ is a known target signature and $\bm{x}$ is a data sample. The background is assumed to be a Gaussian distribution parametrized by $\bm{\mu}$ and $\bm{\Sigma}$ which represent the mean and covariance, respectively. 

\begin{figure}[thb]
				\centering
				\includegraphics[width=0.5\textwidth]{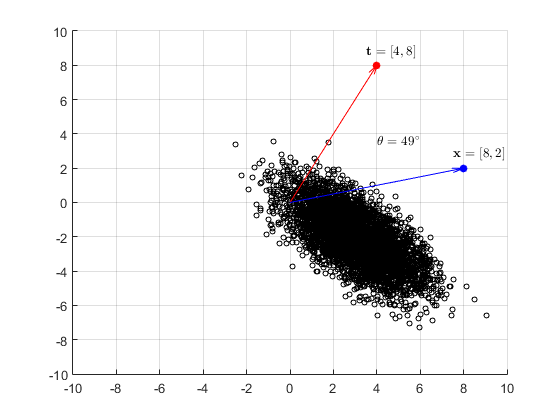}
				\caption{Original example target signature $\bm{t}$, data sample $\bm{x}$, and randomly drawn background samples from a Gaussian, parametrized by $\bm{\mu}$ and $\bm{\Sigma}$. The relative angle between the two vectors is also shown.}
				\label{org_fig}
			\end{figure}

\begin{figure}[thb]
				\centering
				\includegraphics[width=0.5\textwidth]{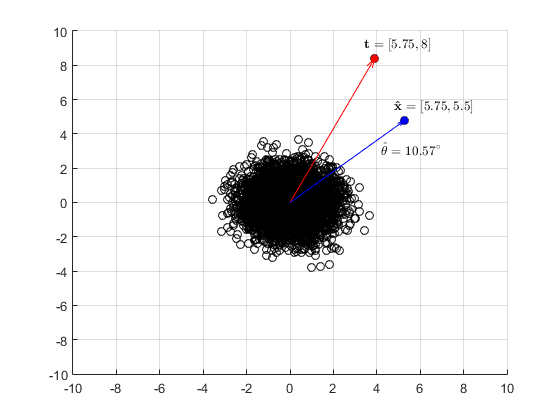}
				\caption{Transformed example target, data, and background vectors. The mean is subtracted and the inverse covariance matrix to the one half power is multiplied to transform each vector. In this case, the angle between $\bm{t}$ and $\bm{x}$ is decreased resulting in a higher ACE statistic.}
				\label{whit_fig}
			\end{figure}

The ACE statistic is a number between zero and one which can be interpreted as a measurement of the presence of $\bm{t}$ in $\bm{x}$. Geometrically, ACE can be estimated as the square of the cosine of the angle between $\bm{x}$ and $\bm{t}$, in a coordinate space transformed by the background estimation. This is illustrated in Fig. \ref{org_fig} and Fig. \ref{whit_fig}, where the point cloud represents samples from the background model. A target vector is represented as $\bm{t}$ and a data sample is represented by $\bm{x}$. 		
In this example, ACE produces 0.84, indicating a relatively strong presence of $\bm{t}$ in $\bm{x}$. The key to effective ACE performance is accurate background estimation. Two approaches for estimation of background mean and covariance are described in the following two sub-sections. 
	
	\subsection{Global ACE}	
Appropriate background estimation is very important for ACE's performance as a detection statistic. ACE assumes that background points can be modeled as a multivariate random Gaussian distribution, parametrized by a mean and covariance. In the simplest case, an entire data set or representative subset is processed at once to generate an estimate of the mean and covariance. This estimate is then used for the entire data set or subset that ACE is computed on. This method is referred to as Global ACE. While this method is non-causal, it is relatively quick to execute because the background model parameters are only computed once. Global ACE also produces comparable results to more complicated background estimations, in our experiments. This makes it useful for testing other aspects of our detection algorithm, such as clustering or pre-processing. Once the background model is estimated, the ACE statistic is computed using each element in the dictionary. The maximum ACE statistic across all the elements is stored as the point’s final confidence.

	 \subsection{Woodbury Identity Updated ACE}
The second method is a causal background estimation approach and takes advantage of the fact that only the inverse of the covariance matrix is actually used in ACE. This method, referred to as the Woodbury Identity Updated Adaptive Coherence Estimator (WACE), uses the Woodbury Inverse Matrix Identity \cite{WOOD_IDEN} to update the inverse covariance matrix directly. This identity has previously been used for background estimation in anomaly detection using a ground penetrating radar system. \cite{WOOD_2}. The mean and covariance estimates are updated using a lagging window running total style estimation. For the WEMI data sets used in our experiments, the first $N$ samples can be assumed to contain mostly background points. These first N samples are used to initialize the background model parameters. This initial estimate is used to compute the ACE statistic for the first 2N samples. From then on, the $k-N^{th}$ ACE statistic is compared with a threshold, where k is the current sample index. If the value is below the threshold, the point is considered part of the background and thus is added to the estimates using Equations \ref{mean_update_eq} and \ref{inv_cov_update_eq}, 
			\begin{equation} \label{mean_update_eq}
				\bm{\mu}_{k+1} = (1 - \lambda)\bm{\mu}_k + \lambda(\bm{x}_{k-N} - \bm{\mu}_k)
			\end{equation}
			\begin{equation} \label{inv_cov_update_eq}
				\bm{\Sigma}_{k+1}^{-1} = \frac{1}{1 - \lambda}\left[\bm{\Sigma}_k^{-1}-\frac{(\bm{x}_{k-N}-\bm{\mu}_k)(\bm{x}_{k-N}-\bm{\mu}_k)^T}{\frac{1-\lambda}{\lambda}+(\bm{x}_{k-N}-\bm{\mu}_k)^T(\bm{x}_{k-N}-\bm{\mu}_k)} \right]
			\end{equation}
where $\bm{\Sigma}^{-1}$ is the inverse covariance estimate, $\bm{x}$ is a data sample, k is the sample index and $\lambda$ is a small scalar between zero and one that controls how much value to place on new background samples. Increasing $\lambda$ will make the estimate more adaptive to new background points. For our experiments, $\lambda$ was set to 0.005.

\section{Experimental Results} \label{results_lbl}
	\subsection{Data Set Description} \label{data_set_lbl}
Experiments were performed on wideband electromagnetic induction data collected at a test facility. The data set provided consists of fourteen subsets, called lanes. Each subset contains data from a physical lane where explosive hazards were buried. The cart holding the WEMI system is driven down each of these lanes while collecting WEMI data. With each WEMI data sample, a UTM coordinate is recorded from the on-board GPS system and stored in a header file. The data set consists of a wide variety of explosive hazards, as well as metal and non-metal clutter. The targets were buried at various depths and can be classified by their metallic content and purpose. Object purposes include anti-tank (AT) and anti-personnel (AP). The data sets metallic content is summarized in Table \ref{tar_desc_tbl}.
			\begin{table}[thb] \label{tar_desc_tbl}
				\centering
				\caption{Object description for WEMI data. Abbreviations, MT: Metal Target, LMT: Low-Metal Target, NMT: Non-Metal Target, CL: Clutter.}
				\begin{tabular}{c c c c c c}
					\hline\hline
					 & \textbf{MT} & \textbf{LMT} & \textbf{NMT} & \textbf{CL}\\
					\textbf{Lane 1} & 4  & 7  & 0  & 6 \\
					\textbf{Lane 2} & 4  & 10 & 0  & 4 \\
					\textbf{Lane 3} & 4  & 7  & 0  & 8 \\
					\textbf{Lane 4} & 6  & 6  & 3  & 0 \\
					\textbf{Lane 5} & 7  & 5  & 5  & 0 \\
					\textbf{Lane 6} & 6  & 6  & 2  & 3 \\
					\textbf{Totals} & 31 & 41 & 10 & 11 \\
					\hline
				\end{tabular}
			\end{table}	

	\subsection{Alarm Generation} \label{alarm_gen_lbl}
To score the results of ACE and other detectors, alarms must be generated. Alarms are, at minimum, a structure containing a UTM coordinate and a confidence value. To generate alarms, an approach which we call morphological clustering is used. This algorithm first maps all of the ACE confidence values and their respective UTM coordinates to a grid using linear interpolation. The location and value of the maximum confidence value in the grid is used to generate an alarm. After the alarm is generated, all points within a 0.5m halo set to zero. The process is repeated until every point in the map is zero. This method for generation ensures monotonically decreasing confidence values. It also must cover the entire lane, guaranteeing total detection with at least the lowest operating threshold. One disadvantage of this method is that a singular noisy point which resulted in a high ACE confidence will have its own alarm assigned, and its surrounding area zeroed out, potentially removing relatively high confidence points over locations where a targets are actually buried. 

	\subsection{Results}
\begin{figure}[tbh]
				\centering
				\includegraphics[width=0.8\textwidth]{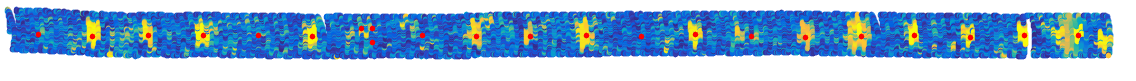}
				\caption{Woodbury Identity Updated Adaptive Coherence Estimator confidence map. High confidence points are colored yellow, low confidence points are colored blue. Red dots are plotted at ground truth centers. Note: Not all ground truth points are targets (i.e., some ground truth points correspond to clutter objects in the lane).}
				\label{conf_map_fig}
			\end{figure}

\begin{figure}[tbh]
				\centering
				\includegraphics[width=0.5\textwidth]{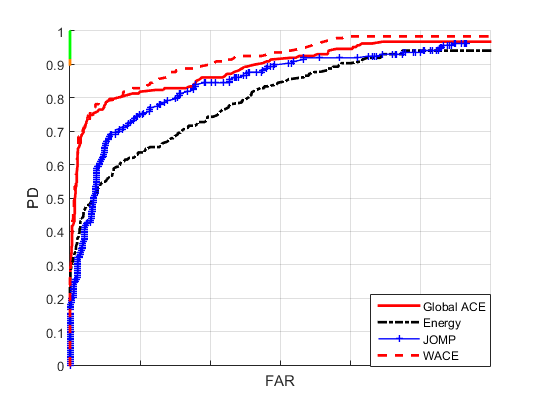}
				\caption{Receiver Operator Curves for each detector applied to the WEMI data set. Alarms over clutter are ignored.}
				\label{roc_fig}
			\end{figure}

\begin{figure}[tbh]
				\centering
				\includegraphics[width=0.5\textwidth]{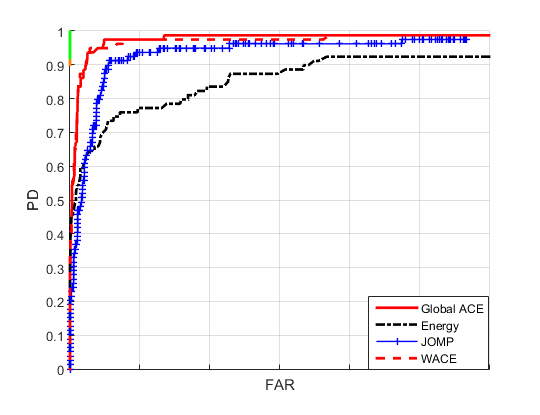}
				\caption{Receiver Operator Curves for each detector applied to the WEMI data set, excluding all but anti-personnel (AP) targets. Alarms over clutter are ignored.}
				\label{roc_ap_only_fig}
			\end{figure}

The receiver operator curve in Figure \ref{roc_fig} shows the results of four implemented detectors applied to the WEMI data described in Section \ref{data_set_lbl}. Each method was applied to the fourteen lanes of WEMI buried explosive data. The Energy method simply uses the sum of the magnitude of the responses at each operating frequency for confidence values. All other detectors use the same one hundred DSRF elements as a target dictionary. The alarm generation method used for all detectors is the described in Section \ref{alarm_gen_lbl}. Confidence maps are often plotted to visualize results for qualitative comparison. An example WACE confidence map for a single lane is shown in Figure \ref{conf_map_fig}.
The UTM coordinate for each alarm is compared to a list of ground truth points to generate a receiver operator curve (ROC). Alarms located within a small halo (0.25m) of targets are considered true positives. Alarms over clutter, both metal and non-metal, were ignored in scoring. Targets at a depth below eight inches were also ignored. All other alarms are considered false alarms. Two ROCs are shown. Figure \ref{roc_fig} shows the scoring results on all targets. Figure \ref{roc_ap_only_fig} shows the scoring results exuding all but anti-personnel (AP) targets.

	\subsection{Conclusion}
All methods show a marked improvement over the Energy detector, which serves as a baseline. JOMP is naturally more conservative, with its two tapped approach when it comes to assigning high confidences. A high JOMP confidence requires matching target signatures in two nearby points instead of just one such as in ACE. While this makes JOMP resilient to random fluctuations which may resemble targets, there is nothing to adjust for background distributions. ACE, however, transforms the data and target vectors according to the background estimate before comparing them. This helps ACE to discriminate between targets and non-targets more easily. WACE preforms similarly to ACE Global, with the benefit of being causal and thus theoretically implementable in real-time. Both ACE based algorithms outperform JOMP and Energy at all false alarm rates. WACE also has the advantage of efficiently updating the inverse covariance matrix, compared to laboriously re-estimating it with each new background point. In addition, the lagging window in WACE is intended to prevent the background estimate from learning current targets. The running total style estimation also makes WACE more adaptive to local background distributions. 
	
	\acknowledgments
	This work was funded by Army Research Office grant number 66398-CS to support the US Army RDECOM CERDEC NVESD. The views and conclusions contained in this document are those of the authors and should not be interpreted as representing the official policies either expressed or implied, of the Army Research Office, Army Research Laboratory, or the U.S. Government. The U.S. Government is authorized to reproduce and distribute reprints for Government purposes notwithstanding any copyright notation hereon.

	\bibliographystyle{./spiebib}
	\bibliography{./refs}
\end{document}